\documentclass[10pt,twocolumn,letterpaper]{article}

\usepackage{cvpr}              
\usepackage{wrapfig}
\usepackage{graphicx}
\usepackage{amsmath}
\usepackage{amssymb}
\usepackage{booktabs}
\usepackage[dvipsnames]{xcolor}
\usepackage{xcolor,colortbl}
\usepackage{color, colortbl}
\usepackage{xcolor}
\newcommand{\tableCellHeight}{1}
\newcommand{\tabstyle}[1]{
  \setlength{\tabcolsep}{#1}
  \renewcommand{\arraystretch}{\tableCellHeight}
  \centering
  \small
}
\usepackage{sidecap}
\newcommand{\tablestyle}[2]{\setlength{\tabcolsep}{#1}\renewcommand{\arraystretch}{#2}\centering\footnotesize}
\usepackage{float}
\definecolor{purple}{RGB}{230, 227, 254}
\definecolor{lightgreen}{RGB}{238, 252, 241}
\definecolor{lightred}{RGB}{231, 187, 187}
\definecolor{darkred}{RGB}{198, 129, 129}

\definecolor{tabhighlight}{HTML}{e5e5e5}

\newcommand{\rotbox}[1]{\rotatebox{55}{#1}}
\usepackage{graphicx, amsmath, amssymb, caption, subcaption, multirow, overpic}
\definecolor{tabhighlight}{HTML}{e5e5e5}
\definecolor{citecolor}{HTML}{0071bc}

\def\ie{\emph{i.e.}\xspace}
\def\eg{\emph{e.g.}\xspace}

\newcommand{\txt}[1]{{\texttt{#1}}}

\usepackage[pagebackref,breaklinks,colorlinks]{hyperref}

\usepackage[capitalize]{cleveref}
\crefname{section}{Sec.}{Secs.}
\Crefname{section}{Section}{Sections}
\Crefname{table}{Table}{Tables}
\crefname{table}{Tab.}{Tabs.}

\def\cvprPaperID{7620} 
\def\confName{CVPR}
\def\confYear{2023}

\begin{document}

\title{{MaPLe}: \textbf{M}ulti-mod\textbf{a}l \textbf{P}rompt \textbf{Le}arning}


 \author{
  Muhammad Uzair Khattak$^{1}$ \quad 
  Hanoona Rasheed$^{1}$ \quad 
  Muhammad Maaz$^{1}$ \\
  Salman Khan$^{1,2}$ \quad
  Fahad Shahbaz Khan$^{1,3}$
  \vspace{0.2em} \\
  $^{1}$Mohamed bin Zayed University of AI \quad 
  $^{2}$Australian National University \quad 
  $^{3}$Link\"{o}ping University
}

\maketitle
\begin{abstract}

Pre-trained vision-language (V-L) models such as CLIP have shown excellent generalization ability to downstream tasks. However, they are sensitive to the choice of input text prompts and require careful selection of prompt templates to perform well. Inspired by the Natural Language Processing (NLP) literature, recent CLIP adaptation approaches learn prompts as the textual inputs to fine-tune CLIP for downstream tasks. We note that using prompting to adapt representations in a single branch of CLIP (language or vision) is sub-optimal since it does not allow the flexibility to dynamically adjust both representation spaces on a downstream task. In this work, we propose Multi-modal Prompt Learning (MaPLe) for \emph{both} vision and language branches to improve alignment between the vision and language representations. Our design promotes strong coupling between the vision-language prompts to ensure mutual synergy and discourages learning independent uni-modal solutions. Further, we learn separate prompts across different early stages to progressively model the stage-wise feature relationships to allow rich context learning. We evaluate the effectiveness of our approach on \emph{three} representative tasks of generalization to novel classes, new target datasets and unseen domain shifts. {Compared with the state-of-the-art method Co-CoOp, MaPLe exhibits favorable performance and achieves an absolute gain of 3.45\% on novel classes and 2.72\% on overall harmonic-mean, averaged over 11 diverse image recognition datasets.} Our code and pre-trained models are available at \href{https://github.com/muzairkhattak/multimodal-prompt-learning}{https://github.com/muzairkhattak/multimodal-prompt-learning}.
\end{abstract}

\section{Introduction}
\label{sec:intro}
\noindent Foundational vision-language (V-L) models such as CLIP (Contrastive Language-Image Pretraining) \cite{radford2021learning} have shown excellent generalization ability to downstream tasks. Such models are trained to align language and vision modalities on web-scale data \eg, 400 million text-image pairs in CLIP. These models can reason about open-vocabulary visual concepts, thanks to the rich supervision provided by natural language. During inference, hand-engineered text prompts are used \eg, ‘\texttt{a photo of a $<$category$>$}’ as a query for text encoder. The output text embeddings are matched with the visual embeddings from an image encoder to predict the output class. Designing high quality contextual prompts have been proven to enhance the performance of CLIP and other V-L models~\cite{jin2021good, yao2021cpt}.

\begin{figure*}[!t]
\centering
{\includegraphics[width=0.9\textwidth]{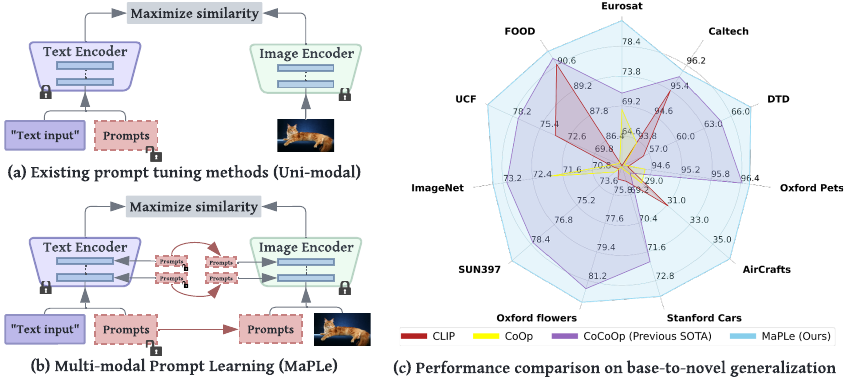}} \caption{\small Comparison of MaPLe with standard prompt learning methods. {\color{blue}\textbf{(a)}} Existing methods adopt uni-modal prompting techniques to fine-tune CLIP representations as prompts are learned only in a single branch of CLIP (language or vision). {\color{blue}\textbf{(b)}} MaPLe introduces branch-aware hierarchical prompts that adapt both language and vision branches simultaneously for improved generalization. {\color{blue}\textbf{(c)}} MaPLe surpasses state-of-the-art methods on 11 diverse image recognition datasets for novel class generalization task.}
\label{fig:intro_figure}
\end{figure*}

    Despite the effectiveness of CLIP towards generalization to new concepts, its massive scale {and scarcity of training data (\eg, few-shot setting)} makes it infeasible to fine-tune the full model for downstream tasks. Such fine-tuning can also forget the useful knowledge acquired in the large-scale pretraining phase and can pose a risk of overfitting to the downstream task. To address the above challenges, existing works propose language prompt learning to avoid manually adjusting the prompt templates and providing a mechanism to adapt the model while keeping the original weights frozen \cite{zhou2022conditional, zhou2022learning, lu2022prompt,huang2022unsupervised, shu2022tpt}. Inspired from Natural Language Processing (NLP), these approaches only explore prompt learning for the text encoder in CLIP (Fig.~\ref{fig:intro_figure}:a) {while adaptation choices together with an equally important image encoder of CLIP remains an unexplored topic in the literature.}

Our motivation derives from the multi-modal nature of CLIP, where a text and image encoder co-exist and \emph{both} contribute towards properly aligning the V-L modalities. {We argue that any prompting technique should adapt the model completely and} therefore, learning prompts only for the text encoder in CLIP is not sufficient to model the adaptations needed for the image encoder. {To this end, we set out to achieve completeness in the prompting approach and} propose \textbf{M}ulti-mod\textbf{a}l \textbf{P}rompt \textbf{Le}arning (MaPLe) to adequately fine-tune the text and image encoder representations such that their optimal alignment can be achieved on the downstream tasks (Fig.~\ref{fig:intro_figure}:b). Our extensive experiments on three key representative settings including base-to-novel generalization,  cross-dataset evaluation, and domain generalization demonstrate the strength of MaPLe. {On base-to-novel generalization, our proposed MaPLe outperforms existing prompt learning approaches across 11 diverse image recognition datasets (Fig. \ref{fig:intro_figure}:c) and achieves absolute average gain of 3.45\% on novel classes and 2.72\% on harmonic-mean over the state-of-the-art method Co-CoOp \cite{zhou2022conditional}. Further, MaPLe demonstrates favorable generalization ability and robustness in cross-dataset transfer and domain generalization settings, leading to consistent improvements compared to existing approaches. Owing to its streamlined architectural design, MaPLe exhibits improved efficiency during both training and inference without much overhead, as compared to Co-CoOp which lacks efficiency due to its image instance conditioned design.} In summary, the main contributions of this work include: \vspace{-0.5em}
\begin{itemize}\setlength{\itemsep}{0em}
\item We propose \emph{multi-modal} prompt learning in CLIP to favourably align its vision-language representations. To the best of our knowledge, this is the first multi-modal prompting approach for fine-tuning CLIP.
\item To link prompts learned in text and image encoders, we propose a \emph{coupling function} to explicitly condition vision prompts on their language counterparts. It acts as a bridge between the two modalities and allows mutual propagation of gradients to promote synergy.
\item Our multi-modal prompts are learned across multiple transformer blocks in both vision and language branches to \emph{progressively} learn the synergistic behaviour of both modalities. This deep prompting strategy allows modeling the contextual relationships independently, thus providing more flexibility to align the vision-language representations. 
\end{itemize}

\begin{figure*}[!t]
\centering
{\includegraphics[scale=0.7]{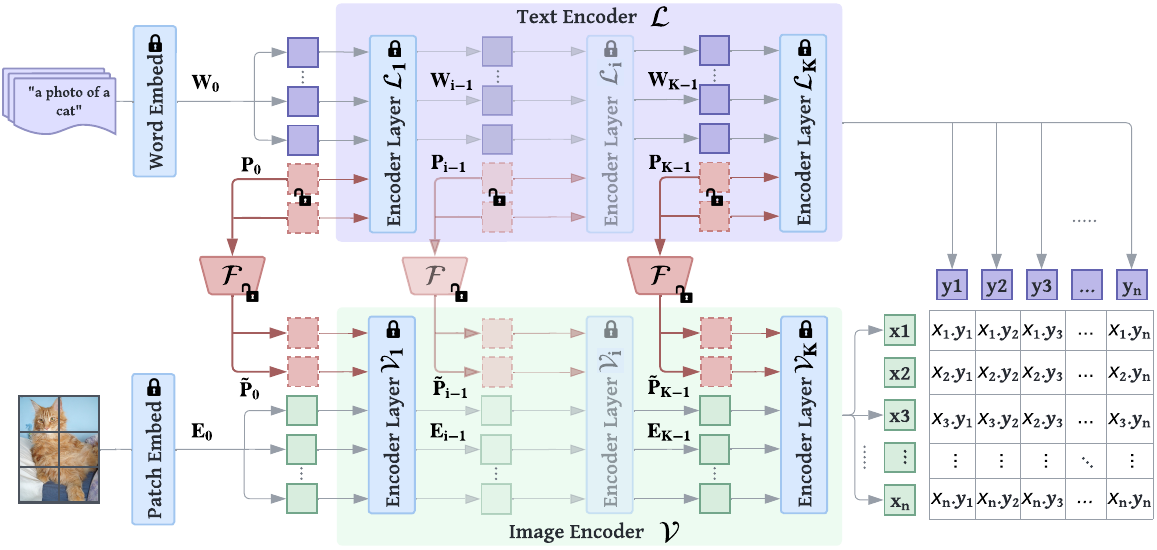}}
\caption{\small Overview of our proposed MaPLe (\textbf{M}ulti-mod\textbf{a}l \textbf{P}rompt \textbf{Le}arning) framework for prompt learning in V-L models. MaPLe tunes both {\setlength{\fboxsep}{0.7pt}\colorbox{lightgreen}{vision}} and {\setlength{\fboxsep}{0.7pt}\colorbox{purple}{language}} branches where only the  {\setlength{\fboxsep}{0.7pt}\colorbox{lightred}{context prompts}} are learned, while the rest of the model is frozen. MaPLe conditions the vision prompts on language prompts via a V-L coupling function $\mathcal{F}$ to induce mutual synergy between the two modalities. Our framework uses deep contextual prompting where separate context prompts are learned across multiple transformer blocks.}
\label{ovd_block_diag}
\end{figure*}

\section{Related Work}
\noindent \textbf{Vision Language Models:}
The combined use of language supervision with natural images is found to be of great interest in the computer vision community. In contrast to models learned with only image supervision, these vision-language (V-L) models encode rich 
multimodal representations. Recently, V-L models like CLIP \cite{radford2021learning}, ALIGN \cite{jia2021scaling}, LiT \cite{zhai2022lit}, FILIP \cite{yao2021filip} and Florence \cite{yuan2021florence} have demonstrated exceptional performance on a wide spectrum of tasks including few-shot and zero-shot visual recognition. These models learn joint image-language representations in a self-supervised manner using abundantly available data from the web.
For example, CLIP and ALIGN respectively use $\sim$400M and $\sim$1B image-text pairs to train a multi-modal network. Although these pre-trained V-L models learn generalized representations, efficiently adapting them to downstream tasks is still a challenging problem. Many works have demonstrated better performance on downstream tasks by using tailored methods to adapt V-L models for few-shot image-recognition \cite{gao2021clip,zhang2021tip,kim2022how}, object detection \cite{rasheed2022bridging, Maaz2022Multimodal, zhou2022detecting, gu2021open, zang2022open, feng2022promptdet}, and segmentation \cite{li2022language,rao2022denseclip,ding2022decoupling,luddecke2022image}. 
In this work, we propose a novel multi-modal prompt learning technique to effectively adapt CLIP for few-shot and zero-shot visual recognition tasks.

\noindent  \textbf{Prompt Learning:}
The instructions in the form of a sentence, known as text prompt, are usually given to the language branch of a V-L model, allowing it to better understand the task. Prompts can be handcrafted for a downstream task or learned automatically during fine-tuning stage. The latter is referred to as ‘Prompt Learning’ which was first used in NLP~\cite{li2021prefix,lester2021power,liu2021p} followed by the adaptation in V-L~\cite{zhou2022learning,zhou2022conditional, zhu2022prompt} and vision-only~\cite{jia2022visual,zhang2022neural,wang2022dualprompt,wang2022learning} models. Similar to \cite{jia2022visual} our
design also uses deep ‘vision’ prompting. However, ours
is the first multi-modal prompting design while \cite{jia2022visual} is uni-modal. 

\noindent  \textbf{Prompt Learning in Vision Language models:}
Full fine-tuning and linear probing~\cite{gao2021clip} are two typical approaches to adapt a V-L model (\ie~CLIP) to the downstream tasks. The complete fine-tuning results in degrading the previously learned joint V-L representation while linear probing limits the zero-shot capability of CLIP.
To this end, inspired from prompt learning in NLP, many works have proposed to adapt V-L models by learning the prompt tokens in an end-to-end training. CoOp~\cite{zhou2022learning} fine-tunes CLIP for few-shot transfer by optimizing continuous set of prompt vectors at its language branch. Co-CoOp~\cite{zhou2022conditional} highlights the inferior performance of CoOp on novel classes and solves the generalization issue by explicitly conditioning prompts on image instances.
\cite{lu2022prompt} proposes to optimize multiple set of prompts by learning the distribution of prompts. \cite{ju2021prompting} adapt CLIP by learning prompts for video understanding tasks.~\cite{bahng2022visual} perform visual prompt tuning on CLIP by prompting on the vision branch. We note that the existing methods follow independent \textit{uni-modal} solutions and learn prompts either in the language or in the vision branch of CLIP, thus adapting CLIP partially. In this paper, we explore an important question: given the multimodal nature of CLIP, is complete prompting (\ie, in both language and vision branches) better suited to adapt CLIP? Our work is the first to answer this question by investigating the effectiveness of multi-modal prompt learning in order to improve alignment between vision and language representations. 

\section{Method}
\noindent Our approach concerns with fine-tuning a pre-trained multi-modal CLIP for better generalization to downstream tasks through context optimization via prompting. Fig.~\ref{ovd_block_diag} shows the overall architecture of our proposed MaPLe (\textbf{M}ulti-mod\textbf{a}l \textbf{P}rompt \textbf{Le}arning) framework.  Unlike previous approaches~\cite{zhou2022learning, zhou2022conditional} which learn context prompts only at the language branch, MaPLe proposes a joint prompting approach where the context prompts are learned in both vision and language branches. Specifically, we append learnable context tokens in the language branch and explicitly condition the vision prompts on the language prompts via a coupling function to establish interaction between them. To learn hierarchical contextual representations, we introduce deep prompting in both branches through separate learnable context prompts across different transformer blocks. During fine-tuning, only the context prompts along with their coupling function are learned while the rest of the model is frozen.
Below, we first outline the pre-trained CLIP architecture and then present our proposed fine-tuning approach.

\begin{figure*}[h!]
\centering
{\includegraphics[width=0.98\textwidth]{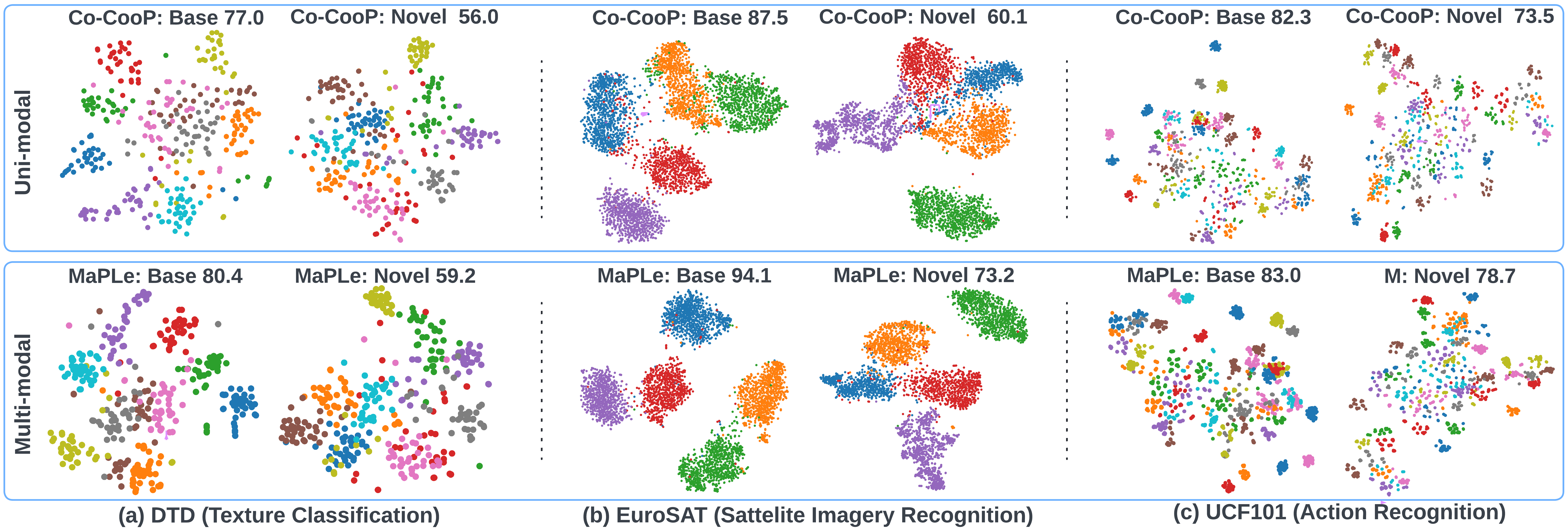}}
\caption{\small t-SNE plots of image embeddings in uni-modal prompting method
Co-CoOp,
and MaPLe on 3 diverse image recognition datasets. MaPLe shows better separability in both base and novel classes.}  
\label{fig:tsne_plots}
\end{figure*}

\subsection{Revisiting CLIP}
\noindent We build our approach on a
pre-trained vision-language (V-L) model, CLIP, which consists of a text and vision encoder. Consistent with existing prompting methods \cite{zhou2022learning, zhou2022conditional}, we use a vision transformer (ViT) \cite{dosovitskiy2020image} based CLIP model. CLIP encodes an image $I \in \mathbb{R}^{H\times W \times 3}$ and a corresponding text description as explained below.

\noindent \textbf{Encoding Image:}
Image encoder $\mathcal{V}$ with $K$ transformer layers $\{\mathcal{V}_i\}_{i=1}^{K}$, splits the image $I$ into $M$ fixed-size patches which are projected into patch embeddings $E_0 \in \mathbb{R}^{M \times d_v}$.
Patch embeddings $E_i$ are input to the $(i+1)^{\text{th}}$ transformer block $(\mathcal{V}_{i+1})$ along with a learnable class (CLS) token $\text{c}_i$ and sequentially processed through $K$ transformer blocks,
\begin{align*}
[{\text{c}}_i, E_i] &= \mathcal{V}_{i}([\text{c}_{i-1}, E_{i-1}])  ~~~~ i= 1, 2, \cdots, K.
\end{align*}
To obtain the final image representation $x$, the class token $\text{c}_{K}$ of last transformer layer $(\mathcal{V}_{K})$ is projected to a common V-L latent embedding space via \texttt{ImageProj}, 
\begin{align*}
x &= \texttt{ImageProj}({\text{c}}_{K}) \ \ \ \ \ ~~~~ x \in \mathbb{R}^{d_{vl}}.
\end{align*}
\textbf{Encoding Text:} CLIP text encoder generates feature representations for text description by tokenizing the words and projecting them to  word embeddings 
${W}_0 =[w_{0}^{1}, w_{0}^{2}, \cdots, w_{0}^{N}] \in \mathbb{R}^{N \times d_{l}}$.
At each stage, ${W}_i$ is input to the $(i+1)^{\text{th}}$ transformer layer of text encoding branch ($\mathcal{L}_{i+1})$,
\begin{align*}
[{W}_i] = \mathcal{L}_{i}({W}_{i-1}) \ \ \ ~~~~ i= 1, 2, \cdots, K
. \end{align*}
The final text representation $z$ is obtained by projecting the text embeddings corresponding to the last token of the last transformer block $\mathcal{L}_K$ to a common V-L latent embedding space via \texttt{TextProj},
\begin{align*}
z = \texttt{TextProj}({w_{K}^{N}}) \ \ \ \ \ ~~~~ z \in \mathbb{R}^{d_{vl}}.
\end{align*} 
\textbf{Zero-shot Classification:}
For zero-shot classification, text prompts are hand-crafted with class labels $y\in \{1,2,\ldots C\}$ (\eg, ‘\texttt{a photo of a $<$category$>$’}) having $C$ classes. Prediction $\hat{y}$ corresponding to the image $I$ having the highest cosine similarity score ($sim(\cdot)$) is calculated with a temperature parameter $\tau$,
\begin{align*}
p(\hat{y}|x) = \frac{\text{exp}(sim(x, z_{\hat{y}})/\tau)}{\sum_{i=1}^{C}\text{exp}(sim(x, z_{i}))}.
\end{align*}
%

\subsection{MaPLe: \textbf{M}ulti-mod\textbf{a}l \textbf{P}rompt \textbf{Le}arning}
\noindent To efficiently fine-tune CLIP for downstream image recognition tasks, we explore the potential of \textit{multi-modal} prompt tuning.
We reason that prior works that have predominantly explored uni-modal approaches are less suitable as they do not offer the flexibility to dynamically adapt both language and vision representation spaces. Thus to achieve completeness in prompting, we underline the importance of multi-modal prompting approach. In Fig.~\ref{fig:tsne_plots}, we visualize and compare the image embeddings of MaPLe with recent state-of-the-art work, Co-CoOp. Note that the image embeddings of CLIP, CoOp and Co-CoOp will be identical as they do not learn prompts in the vision branch. The visualization shows that image embeddings of MaPLe are more separable indicating that learning vision prompts in addition to language prompts leads to better adaptation of CLIP.

In addition to multi-modal prompting, we find that it is essential to learn prompts in the deeper transformer layers to progressively model stage-wise feature representations. To this end, we propose to introduce learnable tokens in the first $J$ (where $J < K$) layers of both vision and language branches. These multi-modal hierarchical prompts utilize the knowledge embedded in CLIP model to effectively learn task relevant contextual representations (see Fig.~\ref{fig:ablation1}).

\subsubsection{Deep Language Prompting}
To learn the language context prompts, we introduce $b$ learnable tokens $\{P^i \in \mathbb{R}^{d_l}\}_{i=1}^b$, in the language branch of CLIP. The input embeddings now follow the form $[P^1, P^2, \cdots, P^b, W_{0}]$, where $W_{0}=[w^{1}, w^{2}, \cdots, w^{N}]$ corresponds to fixed input tokens. New learnable tokens are further introduced in each transformer block of the language encoder ($\mathcal{L}_{i}$) up to a specific depth $J$,
\begin{align}
\label{eq:1}
[\ \underline{\hspace{0.3cm}}, \ W_{i}] = \mathcal{L}_{i}([P_{i-1}, W_{i-1}]) ~~~~ i=1, 2, \cdots, J.
\end{align}
Here $[\cdot,~\cdot]$  refers to the concatenation operation. After $J^{\text{th}}$ transformer layer, the subsequent layers process previous layer prompts and final text representation $z$ is computed,
\begin{align}
\label{eq:2}
[P_{j}, \ W_{j}] &= \mathcal{L}_{j}([P_{j-1}, W_{j-1}]) ~~~~ j=J+1, \cdots, K, \\
z &= \texttt{TextProj}(w_{K}^{N}). \label{eq:3}
\end{align}
When $J=1$, the learnable tokens $P$ are only applied at the input of first transformer layer, and this deep language prompting technique degenerates to CoOp \cite{zhou2022learning}. 

\subsubsection{Deep Vision Prompting}
Similar to deep language prompting, we introduce $b$ learnable tokens $\{\tilde{P}^i \in \mathbb{R}^{d_v}\}_{i=1}^{b}$, in the vision branch of CLIP alongside the input image tokens. New learnable tokens are further introduced in deeper transformer layers of the image encoder ($\mathcal{V}$) up to depth $J$.
\begin{align*}
[c_i, E_i, \ \underline{\hspace{0.3cm}} \ ] &= \mathcal{V}_{i}([c_{i-1}, E_{i-1}, \tilde{P}_{i-1}])  \ ~~~ i=1, 2, \cdots, J, \\
[c_j, E_j, \tilde{P}_{j}] &= \mathcal{V}_{j}([c_{j-1}, E_{j-1}, \tilde{P}_{j-1}])  ~~~  j=J+1, \cdots, K,\\
x &= \texttt{ImageProj}({c}_K).
\end{align*}
Our deep prompting provides the flexibility to learn prompts across different feature hierarchies within the ViT architecture. We find that sharing prompts across stages is better compared to independent prompts as features are more correlated due to successive transformer block processing. Thus, the later stages do not provide independently-learned complimentary prompts as compared to the early stages.

\subsubsection{Vision Language Prompt Coupling} \label{section:compound_prompt_derivation}
We reason that in prompt tuning it is essential to take a multi-modal approach and \textit{simultaneously} adapt both the vision and language branch of CLIP in order to achieve completeness in context optimization. 
A simple approach would be to naively combine deep vision and language prompting, where both the language prompts $P$, and the vision prompts $\tilde{P}$, will be learned during the same training schedule. We name this design as \textit{`Independent V-L Prompting'}. Although this approach satisfies the requirement of completeness in prompting, this design lacks synergy between vision and language branch as both branches do not interact while learning the task relevant context prompts. 

To this end, we propose a branch-aware multi-modal prompting which tunes vision and language branch of CLIP together by sharing prompts across both modalities. Language prompt tokens are introduced in the language branch up to $J^{\text{th}}$ transformer block similar to deep language prompting as illustrated in Eqs.~\ref{eq:1}-\ref{eq:3}. To ensure mutual synergy between V-L prompts, vision prompts $\tilde{P}$, are obtained by projecting language prompts $P$ via vision-to-language projection which we refer to as \textit{V-L coupling function} $\mathcal{F(\cdot)}$, such that $\tilde{P}_k= {\mathcal{F}_k(P_k)}$. The coupling function is implemented as a linear layer which maps $d_l$ dimensional inputs to $d_v$. This acts as a bridge between the two modalities, thus encouraging mutual propagation of gradients.
\begin{align*}
[c_i, E_i, \ \underline{\hspace{0.3cm}} \ ] &= \mathcal{V}_{i}([c_{i-1}, E_{i-1},\boldsymbol{\mathcal{F}_{i-1}(P_{i-1})}])  \ ~ i=1, \cdots, J \\
[c_j, E_j, \tilde{P}_{j}] &= \mathcal{V}_{j}([c_{j-1}, E_{j-1}, \tilde{P}_{j-1}])  \ \ ~  j=J+1, \cdots, K\\
x &= \texttt{ImageProj}({c}_K)
\end{align*}
Unlike independent V-L prompting, explicit conditioning of $\tilde{P}$ on $P$ helps learn prompts in a shared embedding space between the two branches, thus improving mutual synergy. 

\section{Experiments}
\subsection{Benchmark setting}

\noindent \textbf{Generalization from Base-to-Novel Classes:} We evaluate the generalizability of MaPLe, and follow a zero-shot setting where the datasets are split into base and novel classes. The model is trained only on the base classes in a few-shot setting and evaluated on base and novel categories.

\noindent  \textbf{Cross-dataset Evaluation:} To validate the potential of our approach in cross-dataset transfer, we evaluate  our ImageNet trained model directly on other datasets. Consistent with Co-CoOp, our model is trained on all 1000 ImageNet classes in a few-shot manner.

\noindent  \textbf{Domain Generalization:} We evaluate the robustness of our method on out-of-distribution datasets. Similar to cross-dataset evaluation, we test our ImageNet trained model directly on four other ImageNet datasets that contain various types of domain shifts.

\noindent \textbf{Datasets:}
For generalization from base-to-novel classes and cross-dataset evaluation, we follow \cite{zhou2022learning,zhou2022conditional} and evaluate the performance of our method on 11 image classification datasets which covers a wide range of recognition tasks. This includes two generic-objects datasets, ImageNet~\cite{deng2009imagenet} and Caltech101~\cite{fei2004learning}; five fine-grained datasets,  OxfordPets~\cite{parkhi2012cats}, StanfordCars~\cite{krause20133d}, Flowers102~\cite{nilsback2008automated}, Food101~\cite{bossard2014food}, and FGVCAircraft~\cite{maji2013fine}; a scene recognition dataset SUN397~\cite{xiao2010sun}; an action recognition dataset UCF101~\cite{soomro2012ucf101}; a texture dataset  DTD~\cite{cimpoi2014describing} and a satellite-image dataset EuroSAT~\cite{helber2019eurosat}. For domain generalization, we use ImageNet as source dataset and its four variants as target datasets including ImageNetV2~\cite{recht2019imagenet},  ImageNet-Sketch~\cite{wang2019learning}, ImageNet-A~\cite{hendrycks2021natural} and ImageNet-R~\cite{hendrycks2021many}. 

\noindent \textbf{Implementation Details} We use a few-shot training strategy in all experiments at 16 shots which are randomly sampled for each class. We apply prompt tuning on a pre-trained ViT-B/16 CLIP model where $d_{l}=512$, $d_{v}=768$ and $d_{vl}=512$. For MaPLe, we set prompt depth $J$ to 9 and the language and vision prompt lengths to 2. All models are trained for 5 epochs with a batch-size of 4 and a learning rate of 0.0035 via SGD optimizer on a single NVIDIA A100 GPU.
We report base and novel class accuracies and their harmonic mean (HM) averaged over 3 runs. We initialize the language prompts of the first layer $P_{0}$ with the pre-trained CLIP word embeddings of the template `\txt{a photo of a $<$category$>$}', while for the subsequent layers they are randomly initialized from a normal distribution. For training MaPLe on all 1000 classes of ImageNet as a source model, prompt depth $J$ is set to 3 and the model trained for 2 epochs with learning rate of 0.0026. Hyper-parameters for deep language prompting, deep vision prompting, and independent V-L prompting are detailed in Appendix \ref{appendix:iml_details}. The hyper-parameters are fixed across all datasets.

\begin{table}[t!]
    \centering
    \setlength{\tabcolsep}{1mm}{
    \resizebox{1\linewidth}{!}{
    \begin{tabular}{l cccc}
    \toprule
    Method  & Base Acc. & Novel Acc. & HM & GFLOPS\\
            \midrule
    1: MaPLe shallow ($J=1$) & 80.10 & 73.52 & 76.67  & 167.1\\
    2: Deep vision prompting & 80.24 & 73.43 & 76.68  & 18.0 \\
    3: Deep language prompting &	81.72 & 73.81 &	77.56 & 166.8\\
    4: Independent V-L prompting & 82.15 & 74.07 & 77.90  & 167.0\\
    \midrule
    \rowcolor{tabhighlight}
    5: MaPLe (Ours)  &	\textbf{{82.28}} &	\textbf{75.14} &	\textbf{78.55} & 167.0\\
    \bottomrule
    \end{tabular}
    }}
    \caption{ Comparison of MaPLe with different prompting designs in base-to-novel generalization. Results are averaged over 11 datasets. HM refers to harmonic mean. 
    }
    \label{table:different_V-L_prompting}
\end{table}


\subsection{Prompting CLIP via Vision-Language Prompts}
\noindent \textbf{Prompting Variants:} We first evaluate the performance of different possible prompting design choices as an ablation for our proposed branch-aware multi-modal prompting, MaPLe. These variants include shallow MaPLe, deep language prompting, deep vision prompting and independent V-L prompting. In Table~\ref{table:different_V-L_prompting}, we present the results averaged over the 11 image recognition datasets. Shallow MaPLe (row-1) provides consistant improvements over CoOp and Co-CoOp in terms of generalization. Deep language prompting (row-3) shows improvements over deep vision prompting (row-2), indicating that prompts learned at the language branch provide better adaptation of CLIP. Although separately combining the above two approaches (row-4) further improves the performance, it struggles to achieve comprehensive benefits from the language and vision branches. We hypothesize that this is due to the lack of synergy between the learned vision and language prompts as they do not interact with each other during training. Meanwhile, MaPLe tied with deep prompting (row-4) combines the benefits of prompting in both branches by enforcing interactions through explicit conditioning of vision prompts on the language prompts. It provides improvements on novel and base class accuracies which leads to the best HM of 78.55\%. We explore other possible design choices and present the ablations in Appendix~\ref{appendix:alternate_design_choices}.

\subsection{Base-to-Novel Generalization}
\label{lab:comparision_base_to_new}
\noindent \textbf{Generalization to Unseen Classes:} Table~\ref{table:comparision_with_cocoop} presents the performance of MaPLe in base-to-novel generalization setting on 11 recognition datasets. We compare its performance with CLIP zero-shot, and recent prompt learning works including CoOp~\cite{zhou2022learning} and Co-CoOp~\cite{zhou2022conditional}. In case of CLIP, we use hand-crafted prompts that are specifically designed for each dataset.

In comparison with the state-of-the-art Co-CoOp, MaPLe shows improved performance on both base and novel categories on all 11 datasets with an exception of marginal reduction on only the base class performance of Caltech101. With mutual synergy from the branch-aware multi-modal prompting, MaPLe better generalizes to novel categories on all 11 datasets in comparison with Co-CoOp, and obtains an overall gain from 71.69\% to 75.14\%. When taking into account both the base and novel classes, MaPLe shows an absolute average gain of 2.72\% over Co-CoOp. 

In comparison with CLIP on novel classes, Co-CoOp improves only on 4/11 datasets dropping the average novel accuracy from 74.22\% to 71.69\%. MaPLe is a strong competitor which improves accuracy over CLIP on novel classes on 6/11 datasets, with an average gain from 74.22\% to 75.14\%.

\noindent \textbf{Generalization and Performance on Base Classes:} 
Co-CoOp solves the poor generalization problem in CoOp by conditioning prompts on image instances and shows significant gains in novel categories. However on base classes, it improves over CoOp only on 3/11 datasets with an average drop in performance from 82.69\% to 80.47\%. Meanwhile, the completeness in prompting helps MaPLe improve over CoOp on base classes in 6/11 datasets maintaining the average base accuracy to around 82.28\%, in addition to its improvement in generalization to novel classes.

We find that the training strategies of Co-CoOp can be used to substantially boost the generalization performance of vanilla CoOp (6.8\% gain in novel classes). We therefore compare our method with CoOp$^{\dag}$, which trains CoOp in Co-CoOp setting (refer to Appendix~\ref{appendix:iml_details} for more details).

\begin{table}[!h]
\vspace{-0.1in}
    \centering
    \setlength{\tabcolsep}{4.5mm}{
    \resizebox{0.8\linewidth}{!}{
    \begin{tabular}{l cc|c}
    \toprule
    & Base & Novel & HM \\
    \midrule
    CoOp & \textbf{82.69} & 63.22 & 71.66 \\
    Co-CoOp & 80.47 & 71.69 & 75.83 \\
    \midrule
    CoOp\dag & 80.85 & 70.02 & 75.04 \\
    \rowcolor{tabhighlight}
    MaPLe & 82.28 & \textbf{75.14} & \textbf{78.55} \\
    \bottomrule
    \end{tabular}
    }}
    \caption{ Generalization comparison of MaPLe with CoOp\dag.}
    \label{tab:comparison_with_coop_dag}
    \vspace{-0.15in}
\end{table}

Compare to CoOp$^{\dag}$, the vanilla CoOp model seems to overfit on base classes. When compared to CoOp$^{\dag}$ which attains an average base accuracy of 80.85\%, MaPLe shows an improvement of 1.43\% with the average base accuracy of 82.28\% (Table~\ref{tab:comparison_with_coop_dag}).
 
\begin{table*}[t!]
\tablestyle{6pt}{0}
\addtolength{\tabcolsep}{-6pt}
    \tabstyle{1.5pt}
    \setlength{\tabcolsep}{6pt}
    \begin{subtable}[t]{.32\textwidth}
    \centering
    \caption{\textbf{Average over 11 datasets}}
    \begin{tabular}{l cc|c}
    \toprule
    & Base & Novel & HM \\
    \midrule
    CLIP & 69.34 & 74.22 & 71.70 \\
    CoOp & \textbf{82.69} & 63.22 & 71.66 \\
    Co-CoOp & 80.47 & 71.69 & 75.83 \\
    \midrule
    \rowcolor{tabhighlight}
    MaPLe & 82.28 & \textbf{75.14} & \textbf{78.55} \\
     &  \textcolor{MidnightBlue}{{+1.81}} &  \textcolor{MidnightBlue}{{+3.45}} &  \textcolor{MidnightBlue}{{+2.72}} \\
    \bottomrule
    \end{tabular}
    \end{subtable}
    \vspace{1em}
    \begin{subtable}[t]{.32\textwidth}
    \centering
    \caption{ImageNet.}
    \begin{tabular}{l cc|c}
    \toprule
    & Base & Novel & HM \\
    \midrule
    CLIP & 72.43 & 68.14 & 70.22 \\
    CoOp & {76.47} & 67.88 & 71.92\\
    Co-CoOp & 75.98 & {70.43} & {73.10} \\
    \midrule
    \rowcolor{tabhighlight}
    MaPLe & \textbf{76.66} & \textbf{70.54} & \textbf{73.47} \\
      &  \textcolor{MidnightBlue}{{+0.68}} &  \textcolor{MidnightBlue}{{+0.11}} &  \textcolor{MidnightBlue}{{+0.37}} \\
    \bottomrule
    \end{tabular}
    \end{subtable}
    ~
    \begin{subtable}[t]{.32\textwidth}
    \centering
    \caption{Caltech101}
    \begin{tabular}{l cc|c}
    \toprule
    & Base & Novel & HM \\
    \midrule
    CLIP & 96.84 & {94.00} & 95.40 \\
    CoOp & \textbf{98.00} & 89.81 & 93.73 \\
    Co-CoOp & 97.96 & 93.81 & {95.84} \\
    \midrule
    \rowcolor{tabhighlight}
    MaPLe & 97.74 & \textbf{94.36} & \textbf{96.02} \\
      &  \textcolor{Bittersweet}{{-0.22}} &  \textcolor{MidnightBlue}{{+0.55}} &  \textcolor{MidnightBlue}{{+0.18}} \\
    \bottomrule
    \end{tabular}
    \end{subtable}
    ~
    \begin{subtable}[t]{.32\textwidth}
    \centering
    \caption{OxfordPets}
    \begin{tabular}{l cc|c}
    \toprule
    & Base & Novel & HM \\
    \midrule
    CLIP & 91.17 & 97.26 & 94.12 \\
    CoOp & 93.67 & 95.29 & 94.47 \\
    Co-CoOp & {95.20} & {97.69} & {96.43} \\
    \midrule
        \rowcolor{tabhighlight}
    MaPLe & \textbf{95.43} & \textbf{97.76} & \textbf{96.58} \\
     &  \textcolor{MidnightBlue}{{+0.23}} &  \textcolor{MidnightBlue}{{+0.07}} &  \textcolor{MidnightBlue}{{+0.15}} \\
    \bottomrule
    \end{tabular}
    \end{subtable}
    \vspace{1em}
    \begin{subtable}[t]{.32\textwidth}
    \centering
    \caption{StanfordCars}
    \begin{tabular}{l cc|c}
    \toprule
    & Base & Novel & HM \\
    \midrule
    CLIP & 63.37 & \textbf{74.89} & 68.65 \\
    CoOp & \textbf{78.12} & 60.40 & 68.13 \\
    Co-CoOp & 70.49 & 73.59 & {72.01} \\
    \midrule
        \rowcolor{tabhighlight}
    MaPLe & 72.94 & 74.00 & \textbf{73.47} \\
    &  \textcolor{MidnightBlue}{{+2.45}} &  \textcolor{MidnightBlue}{{+0.41}} &  \textcolor{MidnightBlue}{{+1.46}} \\
    \bottomrule
    \end{tabular}
    \end{subtable}
    ~
    \begin{subtable}[t]{.32\textwidth}
    \centering
    \caption{Flowers102}
    \begin{tabular}{l cc|c}
    \toprule
    & Base & Novel & HM \\
    \midrule
    CLIP & 72.08 & \textbf{77.80} & 74.83 \\
    CoOp & \textbf{97.60} & 59.67 & 74.06 \\
    Co-CoOp & 94.87 & 71.75 & {81.71} \\
    \midrule
        \rowcolor{tabhighlight}
    MaPLe & 95.92 & 72.46 & \textbf{82.56} \\
     &  \textcolor{MidnightBlue}{{+1.05}} &  \textcolor{MidnightBlue}{{+0.71}} &  \textcolor{MidnightBlue}{{+0.85}} \\
    \bottomrule
    \end{tabular}
    \end{subtable}
    ~
    \begin{subtable}[t]{.32\textwidth}
    \centering
    \caption{Food101}
    \begin{tabular}{l cc|c}
    \toprule
    & Base & Novel & HM \\
    \midrule
    CLIP & 90.10 & 91.22 & 90.66 \\
    CoOp & 88.33 & 82.26 & 85.19 \\
    Co-CoOp & {90.70} & {91.29} & {90.99} \\
    \midrule
        \rowcolor{tabhighlight}
    MaPLe & \textbf{90.71} & \textbf{92.05} & \textbf{91.38} \\
     &  \textcolor{MidnightBlue}{{+0.01}} &  \textcolor{MidnightBlue}{{+0.76}} &  \textcolor{MidnightBlue}{{+0.39}} \\
    \bottomrule
    \end{tabular}
    \end{subtable}
    \vspace{1em}
    \begin{subtable}[t]{.32\textwidth}
    \centering
    \caption{FGVCAircraft}
    \begin{tabular}{l cc|c}
    \toprule
    & Base & Novel & HM \\
    \midrule
    CLIP & 27.19 & \textbf{36.29} & {31.09} \\
    CoOp & \textbf{40.44} & 22.30 & 28.75 \\
    Co-CoOp & 33.41 & 23.71 & 27.74 \\
    \midrule
        \rowcolor{tabhighlight}
    MaPLe & 37.44 & 35.61 & \textbf{36.50} \\
      &  \textcolor{MidnightBlue}{{+4.03}} &  \textcolor{MidnightBlue}{{+11.90}} &  \textcolor{MidnightBlue}{{+8.76}} \\
    \bottomrule
    \end{tabular}
    \end{subtable}
    ~
    \begin{subtable}[t]{.32\textwidth}
    \centering
    \caption{SUN397}
    \begin{tabular}{l cc|c}
    \toprule
    & Base & Novel & HM \\
    \midrule
    CLIP & 69.36 & 75.35 & 72.23 \\
    CoOp & {80.60} & 65.89 & 72.51 \\
    Co-CoOp & 79.74 & {76.86} & {78.27} \\
    \midrule
        \rowcolor{tabhighlight}
    MaPLe & \textbf{80.82} & \textbf{78.70} & \textbf{79.75} \\
      &  \textcolor{MidnightBlue}{{+1.08}} &  \textcolor{MidnightBlue}{{+1.84}} &  \textcolor{MidnightBlue}{{+1.48}} \\
    \bottomrule
    \end{tabular}
    \end{subtable}
    ~
    \begin{subtable}[t]{.32\textwidth}
    \centering
    \caption{DTD}
    \begin{tabular}{l cc|c}
    \toprule
    & Base & Novel & HM \\
    \midrule
    CLIP & 53.24 & \textbf{59.90} & 56.37 \\
    CoOp & {79.44} & 41.18 & 54.24 \\
    Co-CoOp & 77.01 & 56.00 & {64.85} \\
    \midrule
        \rowcolor{tabhighlight}
    MaPLe & \textbf{80.36} & 59.18 & \textbf{68.16} \\
      &  \textcolor{MidnightBlue}{{+3.35}} &  \textcolor{MidnightBlue}{{+3.18}} &  \textcolor{MidnightBlue}{{+3.31}} \\
    \bottomrule
    \end{tabular}
    \end{subtable}
    ~
    \begin{subtable}[t]{.32\textwidth}
    \centering
    \caption{EuroSAT}
    \begin{tabular}{l cc|c}
    \toprule
    & Base & Novel & HM \\
    \midrule
    CLIP & 56.48 & {64.05} & 60.03 \\
    CoOp & {92.19} & 54.74 & 68.69 \\
    Co-CoOp & 87.49 & 60.04 & {71.21} \\
    \midrule
        \rowcolor{tabhighlight}
    MaPLe & \textbf{94.07} & \textbf{73.23} & \textbf{82.35} \\
      &  \textcolor{MidnightBlue}{{+6.58}} &  \textcolor{MidnightBlue}{{+13.19}} &  \textcolor{MidnightBlue}{{+11.14}} \\
    \bottomrule
    \end{tabular}
    \end{subtable}
    ~
    \begin{subtable}[t]{.32\textwidth}
    \centering
    \caption{UCF101}
    \begin{tabular}{l cc|c}
    \toprule
    & Base & Novel & HM \\
    \midrule
    CLIP & 70.53 & {77.50} & 73.85 \\
    CoOp & \textbf{84.69} & 56.05 & 67.46 \\
    Co-CoOp & 82.33 & 73.45 & {77.64} \\
    \midrule
        \rowcolor{tabhighlight}
    MaPLe & 83.00 & \textbf{78.66} & \textbf{80.77} \\
     &  \textcolor{MidnightBlue}{{+0.67}} &  \textcolor{MidnightBlue}{{+5.21}} &  \textcolor{MidnightBlue}{{+3.13}} \\
    \bottomrule
    \end{tabular}
    \end{subtable}
    \caption{\small\textbf{Comparison with state-of-the-art methods on base-to-novel generalization}. MaPLe learns multi-modal prompts and demonstrates strong generalization results over existing methods on 11 recognition datasets. Absolute gains over Co-CoOp are indicated in \textcolor{MidnightBlue}{blue}.}
    \label{table:comparision_with_cocoop}
\end{table*}

\begin{SCtable*}[][!h]
    \tabstyle{4pt}
    \scalebox{0.85}{
    \begin{tabular}{l c ccccccccccc}
    \toprule
    & \textbf{Source} & \multicolumn{11}{c}{\textbf{Target}} \\ \cmidrule(lr){2-2} \cmidrule(lr){3-13}
    & \rotbox{ImageNet} & \rotbox{Caltech101} & \rotbox{OxfordPets} & \rotbox{StanfordCars} & \rotbox{Flowers102} & \rotbox{Food101} & \rotbox{Aircraft} & \rotbox{SUN397} & \rotbox{DTD} & \rotbox{EuroSAT} & \rotbox{UCF101} & \rotbox{\emph{Average}} \\
    \midrule
    CoOp & \textbf{71.51} & 93.70 & 89.14 & 64.51 & 68.71 & 85.30 & 18.47 & 64.15 & 41.92 & {46.39} & 66.55 & 63.88 \\
    Co-CoOp & 71.02 &\textbf{ 94.43} & 90.14 & 65.32 & 71.88 & 86.06 & 22.94 & \textbf{67.36} & 45.73 & 45.37 & 68.21 & 65.74 \\
    \midrule
\rowcolor{tabhighlight} MaPLe & 70.72 & 93.53 & \textbf{90.49} & \textbf{65.57} & \textbf{72.23} & \textbf{86.20} & \textbf{24.74} & 67.01 & \textbf{46.49} & \textbf{48.06} & \textbf{68.69} & \textbf{66.30} \\
    \bottomrule
    \end{tabular}}
        \caption{ Comparison of MaPLe with existing approaches on cross-dataset evaluation. Overall, MaPLe achieves competitive performance providing highest average accuracy, indicating better generalization.
    }
    \label{tab:xd}
\end{SCtable*}

\subsection{Cross-Dataset Evaluation}
\noindent We test the cross-dataset generalization ability of MaPLe by learning multi-modal prompts on all the 1000 ImageNet classes and then transferring it directly on the remaining 10 datasets. Table~\ref{tab:xd} shows the performance comparison between MaPLe, CoOp and Co-CoOp. On the ImageNet source dataset, MaPLe achieves performance comparable to competing approaches but demonstrates a much stronger generalization performance by surpassing CoOp in 9/10 and Co-CoOp in 8/10 datasets. Overall, MaPLe shows competitive performance leading to the highest averaged accuracy of 66.30\%. This suggests that the use of branch-aware V-L prompting in MaPLe facilitates better generalization.
\subsection{Domain Generalization}
\vspace{-0.1in}
\noindent We show that MaPLe generalizes favourably on out-of-distribution datasets as compared to CoOp and Co-CoOp. We evaluate the direct transferability of ImageNet trained model to various out-of-domain datasets, and observe that it consistently improves against all the existing approaches as indicated in~Table~\ref{tab:robustness}. This indicates that utilizing multi-modal branch-aware prompting helps MaPLe in enhancing the generalization and robustness of V-L models like CLIP.

\begin{table}[!t]
    \centering
\tablestyle{6pt}{1.1}
\addtolength{\tabcolsep}{-4.5pt}
    \begin{tabular}{l ccccc}
    \toprule
    & \textbf{Source} & \multicolumn{4}{c}{\textbf{Target}} \\ \cmidrule(lr){2-2} \cmidrule(lr){3-6}
     & ImageNet & ImageNetV2 & ImageNet-S & ImageNet-A & ImageNet-R \\
    \midrule
    CLIP &  66.73 & 60.83 & {46.15} & 47.77 & {73.96} \\
    CoOp &  \textbf{71.51} & \textbf{64.20} & 47.99  & 49.71  & 75.21  \\
    Co-CoOp & 71.02 & {64.07} & 48.75 & 50.63 & 76.18  \\
    \midrule
    \rowcolor{tabhighlight} MaPLe & 70.72 & {64.07} & \textbf{49.15} & \textbf{50.90}  & \textbf{76.98} \\
    \bottomrule
    \end{tabular}
        \caption{ Comparison of MaPLe with existing approaches in domain generalization setting. MaPLe shows consistant improvements on all target datasets.
    }
    \label{tab:robustness}
\end{table}

\subsection{Ablation Experiments}
\begin{figure}[!b]
    \includegraphics[width=\columnwidth]{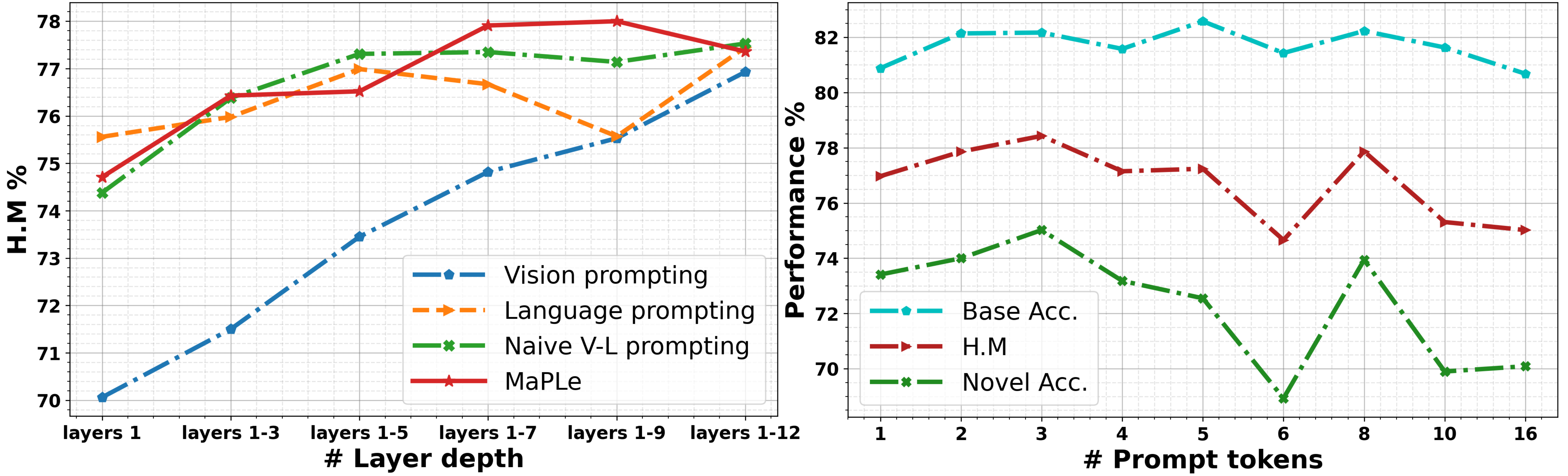}
\caption{\small Ablation on prompt depth  \emph{\textbf{(left)}} and prompt length \emph{\textbf{(right)}} in MaPLe. We report average results on the held-out validation sets of all datasets.}
  \label{fig:ablation1}
\end{figure}

\noindent \textbf{Prompt Depth:} 
\noindent In Fig.~\ref{fig:ablation1} (left), we illustrate the effect of prompt depth $J$ for MaPLe and ablate on the depth of language and vision branch \emph{individually}. In general, the performance improves as prompt depth increases. We note that performance sensitivity increases when randomly initialized prompts are inserted in the deeper layers of a frozen model where the model feature space is already mature. Similar trend is also reported by \cite{jia2022visual}. As earlier methods utilize shallow language prompting ($J=1$), we compare our method with deep language prompting. Overall, MaPLe achieves better performance than deep language prompting and achieves maximum performance at a depth of 9. 

\noindent\textbf{Prompt Length:} 
Fig.~\ref{fig:ablation1} (right) shows the effect of prompt length for MaPLe. As the prompt length increases, the performance on base classes is generally maintained, while the novel class accuracy decreases. This indicates over-fitting which inherently hurts the generalization to novel classes.

\begin{SCfigure}[][!t]
\centering
    \includegraphics[width=0.55\columnwidth]{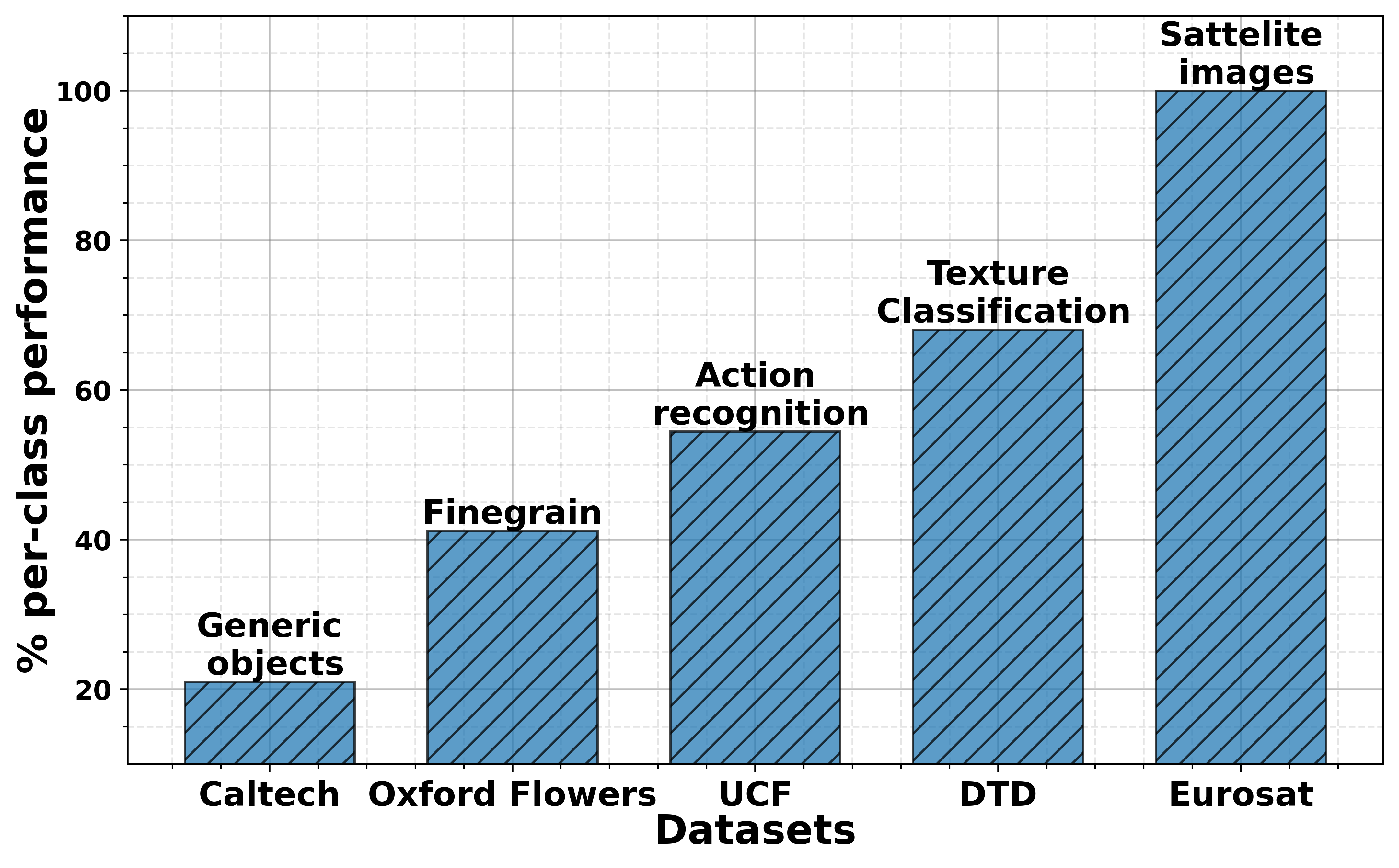}
  \caption{\protect \small Percentage classes where MaPLe shows improved performance over Co-CoOp, which increases as dataset domain shift from generic categories increases ($\rightarrow$).}
  \label{fig:per_class_bar}
  \vspace{-0.1in}
\end{SCfigure}

\noindent\textbf{Effectiveness of Multi-modal Prompting:}
Fig.~\ref{fig:per_class_bar} shows the analysis of per class accuracy for selected datasets in the order of increasing domain shift. It indicates that the performance gains of MaPLe in comparison to Co-CoOp varies across different datasets. MaPLe provides significant gains over Co-CoOp for datasets that have large distribution shifts from the pretraining dataset of CLIP, and vision concepts that are usually rare and less generic. Further detailed analysis is provided in Appendix~\ref{appendix:further_analysis}.

\noindent \textbf{Prompting complexity:}
Table~\ref{tab:computationalcomparison} shows the computational complexity of MaPLe in comparison with other approaches. Although MaPLe utilizes multi-modal prompts, its overall FLOPS (Floating Point Operations) exceeds only by 0.1\% over CoOp and Co-CoOp. The independent V-L prompting also provides comparable FLOP count. In terms of inference speed, Co-CoOp is significantly slower and the FPS (Frames Per Second) remains constant as the batch size increases. In contrast, MaPLe has no such overhead and provides much better inference and training speeds. Further, MaPLe provides better convergence as it requires only half training epochs as compared to Co-CoOp (5 vs 10 epochs). MaPLe adds about 2.85\% training parameters on top of CLIP. To study if the performance gain is mainly attributed to more parameters, we experiment with MaPLe\dag, which uses a unified V-L coupling function for all layer prompts. MaPLe\dag~ with about 9x lesser parameters than MaPLe also improves over existing methods. We also ablate by comparing MaPLe with heavier CoCoOp in Appendix \ref{appendix:heavier_cocoop}.

\begin{table}[!h]
\tablestyle{8pt}{1.1}
\addtolength{\tabcolsep}{-3.7pt}
\scalebox{01}{
\begin{tabular}{lcccccc}
\toprule
    \multirow{2}{*}{Method} & \multirow{2}{*}{Params} &  {Params}   & \multicolumn{3}{c}{ FPS (with BS)} & \multirow{2}{*}{HM}\\ 
    &  & \% CLIP & 1 & 4 & 100 & \\
        \midrule
        CoOp &   2048 & 0.002  & 13.8 & 55.3 & 1353.0 & 71.66\\
        CoCoOp & 35360 & 0.03 & 64.6	&114.7&	15.1 & 75.83 \\
        
        Independent V-L &  31488 & 0.02  & 62.5&	239.4&	1383.8 & 77.90\\
      \rowcolor{tabhighlight}  MaPLe  &  3.55 M & 2.85 &60.2	&239.0&	1365.1 & 78.55\\
      \rowcolor{tabhighlight}  MaPLe\dag & 0.41 M & 0.33 & 60.2&	238.0	&1365.0&   78.11\\
\bottomrule
    \end{tabular}
} 
\caption[caption]{{Comparison of computational complexity among different prompting methods. MaPLe\dag~ is a MaPLe version which utilizes a common V-L coupling function for all layers.}
\label{tab:computationalcomparison}
}
\vspace{-0.15in}
\end{table}

\section{Conclusion}
\noindent Adaptation of large-scale V-L models, \eg, CLIP~\cite{radford2021learning} to downstream tasks is a challenging problem due to the large number of tunable parameters and limited size of downstream datasets. Prompt learning is an efficient and scalable technique to tailor V-L models to novel downstream tasks. To this end, the current prompt learning approaches either consider only the vision or language side prompting.  Our work shows that it is critical to perform prompting for both vision and language branches to appropriately adapt V-L models to downstream tasks. Further, we propose a strategy to ensure synergy between vision-language modalities by explicitly conditioning the vision prompts on textual prompt across different transformer stages.  Our approach improves the generalization towards novel categories, cross-dataset transfer and datasets with domain shifts.

{\small
\bibliographystyle{ieee_fullname}
\bibliography{egbib}
}

\newpage
\newpage
\appendix

\begin{center}
\textbf{\Large Supplementary Material}
\end{center}

 \noindent This section contains supplementary material that provides additional details for the main paper and further experimental analysis. This section follows the contents in the following order.
\begin{itemize}
    \item Additional implementation details (Appendix~\ref{appendix:iml_details})
    \item Alternate prompting design choices (Appendix~\ref{appendix:alternate_design_choices})
    \item Understanding multi-modal prompts (Appendix~\ref{appendix:further_analysis})

    \item Comparison of MaPLe with heavier Co-CoOp (Appendix~\ref{appendix:heavier_cocoop})
\end{itemize}

\section{Additional Implementation details}
 \noindent In this section, we provide further hyper-parameter details of the proposed approaches presented in the main paper. Table \ref{table:hyper_parameter} shows the hyper-parameters chosen for vision, language and independent V-L prompting techniques. We use a learning rate of 0.0025 for language and vision prompting, and 0.0035 for independent V-L prompting.
\label{appendix:iml_details}

\begin{table}[h!]
\tablestyle{6pt}{1.1}
\addtolength{\tabcolsep}{-3pt}
\begin{tabular}{l ccc}
\toprule
Method  & Prompt Depth ($K$) & V-tokens ($\tilde{P}$) & T-tokens ($P$)\\
\midrule
Language prompting &	12 & 0 &	4\\
Vision prompting & 12 & 4 & 0\\
I-V-L prompting & 12 & 2 & 2\\
\bottomrule
\end{tabular}
\caption{
Hyper-parameter settings for deep prompting variants. I-V-L refers to independent V-L prompting. Here $K$ represents the depth of prompts. Number of prompt tokens in vision and language branches are denoted as $\tilde{P}$ and $P$ respectively.
} 
\label{table:hyper_parameter}
\end{table}

\noindent \textbf{CoOp in Co-CoOp setting:} The CoOp approach trained in Co-CoOp setting (denoted by CoOp\dag) uses training configurations of CoCoOp. Similar to Co-CoOp training, CoOp\dag~ trains the standard CoOp method for 10 epochs instead of default 200 epochs. We use a batch size of 4 with a learning rate of 0.0035.

\section{Alternate Design Choices}
\label{appendix:alternate_design_choices}

\noindent \textbf{Prompt Initialization:} 
 Table~\ref{table:prompt_initialization} shows the effect of prompt initialization on MaPLe. Best performance is achieved when the learnable prompts in the first layer are initialized with the prompt `\txt{a photo of a $<$category$>$}' and rest of the layers are initialized randomly (row-3). Initializing prompts with a similar template in all layers leads to lower performance suggesting that 
this is redundant as these prompts learn hierarchically different contextual concepts in different layers (row-1). However, complete random initialization of prompts provides competitive performance (row-2). For implementation, if the number of learnable prompts $M=\#P$ are less than the total tokens of initial prompt template, we convert the former $M$ word embeddings of template with learnable prompts and consider the rest of word embeddings of prompt template as fixed and use all token embeddings (learnable prompts + fixed word tokens) as input to text encoder.

\begin{table}[h!]
\centering
    \tablestyle{6pt}{1.1}
\addtolength{\tabcolsep}{-4.5pt}
\begin{tabular}{l ccc}
\toprule
Method  & Base & Novel & HM\\
\midrule
1: MaPLe: All layers: `\txt{a photo of a}' & 81.90 & 74.22 &	77.88\\
2: MaPLe: Random initialization  & 82.27 &	75.10 &	78.52\\
3: MaPLe: Only first layer: `\txt{a photo of a}' & \textbf{82.28} &	\textbf{75.14} &	\textbf{78.55}\\
\bottomrule
\end{tabular}
\caption{
{Ablation on prompt initialization.} In general, the performance of MaPLe is affected by the choice of prompt initialization.
} 
\label{table:prompt_initialization}
\end{table}

\noindent \textbf{Direction of prompt projection:} As discussed in Section~\ref{section:compound_prompt_derivation}, MaPLe explicitly conditions the vision prompts $\tilde{P}$ on the language prompts $P$ ($P$ $\rightarrow$ $\tilde{P}$) using a V-L coupling function $\mathcal{F}$. Here, we provide analysis for an alternative design choice where $P$ is conditioned on $\tilde{P}$ ($\tilde{P}$ $\rightarrow$ $P$).
Table~\ref{table:ablation_projection_direction} shows that
our approach (\textit{$P$ $\rightarrow$ $\tilde{P}$}) is a better choice which can be reasoned by the lower information loss in such a design since the dimension size $d_{v}$ of vision prompts is greater than the dimension size $d_{l}$ of language prompts.

\begin{table}[t]
\tablestyle{7pt}{1.1}
\addtolength{\tabcolsep}{-1.5pt}
\centering
\begin{tabular}{l ccc}
\toprule
Prompt Proj.  & Base & Novel & HM\\
\midrule
\textbf{$\tilde{P}$ $\rightarrow$ $P$} & 81.37 & 73.25 &	77.10\\
\textbf{$P$ $\rightarrow$ $\tilde{P}$} & \textbf{82.28} &	\textbf{75.14} &	\textbf{78.55}\\
\bottomrule
\end{tabular}
\caption{Projecting from $P$ to $\tilde{P}$ provides the best results.}
\label{table:ablation_projection_direction}
\end{table}

 \noindent \textbf{Exploring other prompting designs:} 
We provide analysis on other possible multi-modal prompting design choices in comparison to MaPLe. As learnable prompts in different transformer layers do not interact with each other, we explore a \textit{progressive prompting} approach where the prompts at each block are conditioned on the prompts from the previous block via linear projection which are then added with the deep prompts initialized at every corresponding layer. We apply this approach to independent V-L prompting (row-1) and MaPLe (row-2). To analyze whether independent V-L prompting and MaPLe provide complementary gains, we also explore a design choice combining them together (row-3) in the same model. The results in Table~\ref{table:other_design_variants_ablation} indicate that MaPLe provides best performance as compared to other design choices.  

\begin{table}[h!]
\tablestyle{6pt}{1.1}
\addtolength{\tabcolsep}{-4.5pt}
\centering
\begin{tabular}{l ccc}
\toprule
Method  & Base & Novel & HM\\
\midrule
1: I-V-L + Progressive prompting &	81.20 & 74.92 &	77.93\\
2: MaPLe + Progressive prompting & 81.45 & 75.04 & 78.11\\
3: MaPLe + I-V-L prompting & 82.27 & 74.05 & 77.94\\
\midrule
\rowcolor{tabhighlight}
4: MaPLe  &	\textbf{82.28} &	\textbf{75.14} &	\textbf{78.55}\\
\bottomrule
\end{tabular}
\caption{
Analysis on alternative design choices for V-L prompting. Overall, MaPLe proves to be the best variant among alternate prompting-related design choices.  
} 
\label{table:other_design_variants_ablation}
\end{table}




\label{appendix:additional_analysis}

\section{Understanding Multi-modal Prompts}
\label{appendix:further_analysis}
\noindent Our experimental results in Section \ref{lab:comparision_base_to_new} indicates that the performance gains of MaPLe in comparison to Co-CoOp varies significantly across different datasets. For some datasets, like ImageNet and Caltech101, the gains are less than 1\%, while
on other datasets like EuroSAT, FGVCAircrafts and DTD, MaPLe shows significant improvements like +13\% over Co-CoOp. To better understand at which cases MaPLe is most effective, we dissect the individual dataset performances and perform an exhaustive per-class analysis. Consistent with earlier work \cite{bahng2022visual}, we conjecture that CLIP pretraining dataset has been curated in a way that maximizes its zero-shot performance on ImageNet-1k and can be used as a proxy for CLIP pretraining dataset. Further, datasets like EuroSAT (satellite images) and DTD (texture dataset) has more distributional gap from ImageNet \cite{bahng2022visual}. Fig.~\ref{fig:per_class_bar} shows per class analysis for selected datasets in the order of increasing diversity (distribution gap w.r.t CLIP pretraining dataset, \ie generic objects).
The overall trend indicates that MaPLe is more effective than Co-CoOp as the diversity of the dataset increases. We conjecture that this is because fine-tuning or prompting bridges the gap between the distribution of the downstream and the pretraining dataset and thus improves the performance. However, the effectiveness would therefore be  less substantial for datasets with little distribution shifts. This intriguing property is also validated for visual prompting in literature~\cite{bahng2022visual}. MaPLe provides completeness in prompting by learning both vision and language prompts to effectively steer CLIP, this makes it more adaptive than Co-CoOp to improve on datasets with larger distribution shifts.

 \noindent Additionally, we note that MaPLe benefits on categories which would have been rarely seen by CLIP during its pretraining (400 million image caption dataset, obtained from internet images). We observe that MaPLe provides significant gains over Co-CoOp for vision concepts that tend to be rare and less generic, \eg, satellite images. In contrast, MaPLe performs competitively to Co-CoOp on frequent and more generic categories \eg, forest, river, dog, \emph{etc}. Multi-modal prompts allow MaPLe to better adapt CLIP for visual concepts that are rarely occurring as compared to existing uni-modal prompting techniques. In Table~\ref{table:class_category_analysis}, we highlight category-wise comparison between MaPLe and Co-CoOp for some selected datasets.

\textbf{Text embeddings analysis:} As all samples within a category are represented using a single text embedding, we take a quantitative approach in Tab. \ref{rebuttal:table_1_cosine_simi} for analyzing the text embeddings of CoOp and MaPLe. We show the pairwise cosine similarity and normalized $l_{2}$ distance metrics averaged across text embeddings. We observe that MaPLe shows better separability among the categories.

\begin{table}[t]
\setlength{\tabcolsep}{3pt}
\centering
\resizebox{0.7\linewidth}{!}{%
\begin{tabular}{lccc|ccc}
\toprule
& \multicolumn{3}{c}{$l_{2}$ distance {$\uparrow$}} & \multicolumn{3}{c}{ Cosine similarity $\downarrow$}  \\
 Method & DTD & UCF & EuroSAT & DTD & UCF & EuroSAT \\
\midrule
CoOp & 0.87 & 0.85 & 0.57  & 0.62 & 0.63 & 0.83 \\
\textbf{MaPLe} & \textbf{0.93}  & \textbf{0.87} &\textbf{ 0.78 }& \textbf{0.57} & \textbf{0.62} & \textbf{0.69}\\
\bottomrule
\end{tabular}}\hspace{-0.9em}
\caption{\small Avg. cosine similarity and $l_2$ distance of text embeddings. MaPLe shows better separability among the text categories.}
\label{rebuttal:table_1_cosine_simi}
\end{table}

\begin{table}[ht]
\tablestyle{8pt}{1.1}
\addtolength{\tabcolsep}{-3.7pt}
\centering
\begin{tabular}{ccc}
\toprule
\multirow{2}{*}{Dataset} & {MaPLe is better}  & {Co-CoOp is better} \\
  & {than  Co-CoOp}  & {than MaPLe} \\
\midrule
Caltech101 & Crontosaurus, & Elephant,\\
(Generic Objects) & Gerenuk, Sea Horse & Ceiling Fan, Cellphone \\
\midrule
EuroSAT & Annual Crop Land, & \multirow{2}{*}{-}\\
(Satellite Image) & Permanent Crop Land &		\\
\midrule
UCF101 & Handstand Walking, & Walking With Dog,\\
(Action recognition) & Playing Daf & Horse Riding	\\
\bottomrule
\end{tabular}
\caption{ 
Analyzing the nature of categories where MaPLe performs better than Co-CoOp. Co-CoOp performs favourably well on generic categories, while MaPLe provides benefits on classes that are typically rare.
} 
\label{table:class_category_analysis}
\end{table}

\section{Comparing MaPLe with Heavier Co-CoOp}
\label{appendix:heavier_cocoop}
 \noindent The multi-modal deep prompting architecture design of MaPLe along with its V-L coupling function $\mathcal{F}$ constitutes more learnable parameters as compared to CoOp and Co-CoOp. 
To verify that the performance gain is not due to increased parameter count, we compare Co-CoOp with MaPLe shallow ($J=1$) that utilizes prompts only at the first layer of vision and language branch of CLIP. Further, we also experiment with a heavier Co-CoOp in which we retrain a version of Co-CoOp that matches the parameter count of MaPLe ($J=9$) by stacking multiple additional layers in its Meta-Net block. Table~\ref{table:heavier_cocoop} indicates the effectiveness of multi-modal prompting in MaPLe (for both $J=1$ and $J=9$) over the heavier Co-CoOp. In addition to that, we experiment with MaPLe\dag, which uses a unified V-L coupling function for all layer prompts. MaPLe\dag~ with about 9x lesser parameters than MaPLe also improves over existing methods. This shows that the difference in the number of parameters is not the cause of gain in our case and the proposed multi-modal prompting design choice makes a difference.

\begin{table}[h]
\tablestyle{6pt}{1.1}
\addtolength{\tabcolsep}{-1.5pt}
\centering
\begin{tabular}{l ccc}
\toprule
Method  & Base& Novel& HM\\
\midrule
Co-CoOp &	80.47 & 71.69 &	75.83\\
Heavier Co-CoOp& 80.14 & 72.02 & 75.86\\
\midrule
\rowcolor{tabhighlight} 
MaPLe shallow ($J=1$) & 80.10 & 73.52 & 76.67 \\
\rowcolor{tabhighlight}  MaPLe\dag~ ($J=9$) & 82.29 &	74.34 & 78.11\\
\rowcolor{tabhighlight} 
MaPLe ($J=9$)  &	\textbf{82.28} &	\textbf{75.14} &	\textbf{78.55}\\
\bottomrule

\end{tabular}
\caption{
Comparison of MaPLe with a heavier Co-CoOp model. We retrain a heavier version of Co-CoOp which is comparable with MaPLe in terms of total parameter count. MaPLe\dag~ is a MaPLe version which utilizes a common V-L coupling function for all layers.
} 
\label{table:heavier_cocoop}
\end{table}

\end{document}